\documentclass[11pt,a4paper]{article}
\usepackage[hyperref]{emnlp2020}
\usepackage{times}
\usepackage{latexsym}

\usepackage{graphicx}
\usepackage{mathtools}
\usepackage{microtype}
\usepackage{multirow}
\usepackage{breqn}
\usepackage{multirow}
\usepackage{booktabs}
\usepackage{url}
\usepackage{tabularx}
\usepackage{subcaption}
\usepackage{bm}
\usepackage{comment}
\usepackage{arydshln}
\usepackage{tabularx}
\usepackage{textcomp}

\usepackage[utf8]{inputenc} 
\usepackage[T1]{fontenc}    
\usepackage{hyperref}       
\usepackage{url}            
\usepackage{booktabs}       
\usepackage{amsfonts}       
\usepackage{nicefrac}       
\usepackage{microtype}      
\usepackage{amsmath}
\usepackage{graphicx}
\usepackage{multirow}
\usepackage{breqn}
\usepackage{bm}
\usepackage{subcaption}
\usepackage{textcomp}
\usepackage{wrapfig}
\usepackage{amssymb}

\usepackage{xcolor}

\newcommand{\dataname}{\textbf{ZUCC}}

\usepackage[textsize=scriptsize]{todonotes}

\aclfinalcopy 
\def\aclpaperid{2574} 

\title{Deriving Commonsense Inference Tasks from Interactive Fictions}

\author{{Mo Yu\thanks{\,\,Equal contribution from the corresponding authors.}$^{\,\,1}$, Xiaoxiao Guo$^{*1}$, Yufei Feng$^{*2}$, Xiaodan Zhu$^{2}$, Michael Greenspan$^{2}$, Murray Campbell$^{1}$}
\\
   $^1$ IBM Research \quad 
   $^2$ Queens University \quad 
   \\
{
    \texttt{yum@us.ibm.com} \quad
    \texttt{xiaoxiao.guo@ibm.com}\quad 
 \texttt{feng.yufei@queensu.ca}
  }}

\begin{document}
\maketitle

\begin{abstract}
Commonsense reasoning simulates the human ability to make presumptions about our physical world, and it is an indispensable cornerstone in building general AI systems. We propose a new commonsense reasoning dataset based on human's interactive fiction game playings as human players demonstrate plentiful and diverse commonsense reasoning. The new dataset mitigates several limitations of the prior art.  Experiments show that our task is solvable to human experts with sufficient commonsense knowledge but poses challenges to existing machine reading models, with a big performance gap of more than 30\%.
\end{abstract}

\section{Introduction}
When playing an Interactive Fiction (IF) game, we explore and progress through a fantasy world by observing textual descriptions and sending text commands to control the protagonist. While in pure texts, we relate the implicit knowledge of these fantasy worlds with those in our physical world. For example, we explore unvisited regions by planning over the mentioned locations (spatial relations); we eat apples to recover health and attach the enemies with swords, but not vice versa (physical interaction relations); we retrospect the poor choice of breaking the lantern when we find the protagonist in a dangerous dark wood (cause and effects). Plentiful and diverse commonsense knowledge from our physical world is encoded in our game playing texts. Unlike most recent approaches focusing on the gameplaying's control problem via model-free Reinforcement Learning (RL), we utilize the IF games to build a new commonsense reasoning dataset. 

There has been a flurry of recent datasets and benchmarks on commonsense reasoning~\cite{levesque2012winograd,zhou2019going,talmor2019commonsenseqa,mullenbach2019nuclear,jiang2020wordcraft,sap2019atomic,bhagavatula2019abductive,huang2019cosmos,bisk2020piqa,sap2019social,zellers2018swag}. 
All these existing benchmarks adopt a multi-choice form task. With the input query and an optional short paragraph of the background description, each candidate forms a statement.  The statement that is consistent with a commonsense knowledge fact corresponds to the correct answer. We notice some common deficiencies in the construction of these benchmarks.
First, nearly all these benchmarks focus on one specific facet and ask human annotators to write candidates related to the specific type of commonsense. As a result, the distribution of these datasets is not natural but biased to a specific facet.
For example, most benchmarks focus on collocation, association or other relations (e.g., ConceptNet~\cite{speer2016conceptnet} relations) between words or concepts~\cite{levesque2012winograd,talmor2019commonsenseqa,mullenbach2019nuclear,jiang2020wordcraft}. Other examples include temporal commonsense~\cite{zhou2019going},
physical interactions between action and objects~\cite{bisk2020piqa}, emotions
and behaviors of people under the given situation~\cite{sap2019social}, and cause-effects between events and states~\cite{sap2019atomic,bhagavatula2019abductive,huang2019cosmos}.
Second, the task form makes them more likely commonsense validation, i.e., validation between a commonsense fact and a text statement, but neglecting hops among multiple facts.\footnote{These datasets do contain a portion of instances that require explicit reasoning capacity, especially~\cite{bhagavatula2019abductive,huang2019cosmos,bisk2020piqa,sap2019social}. Still, many of the instances can be solved with standalone facts.}
The limitations above of previous works, namely 
limitations in \emph{distributions of required commonsense knowledge types} and \emph{forms of tasks}, restricted their potentials. These tasks are naturally easy to be handled with pre-trained Language Models (LMs) such as BERT~\cite{devlin2019bert}. It is mainly because (1) the narrow reasoning types are easier to be fit by a powerful LM; (2) the dominating portion of commonsense validation instances are easier to be captured by pre-training if texts on these facts have presented in pre-training.
Additionally, the above limitations naturally lead to discrepancies between practical NLP tasks that require broad reasoning ability on various facets.

\begin{figure}[t]
\centering
\includegraphics[width=0.95\linewidth]{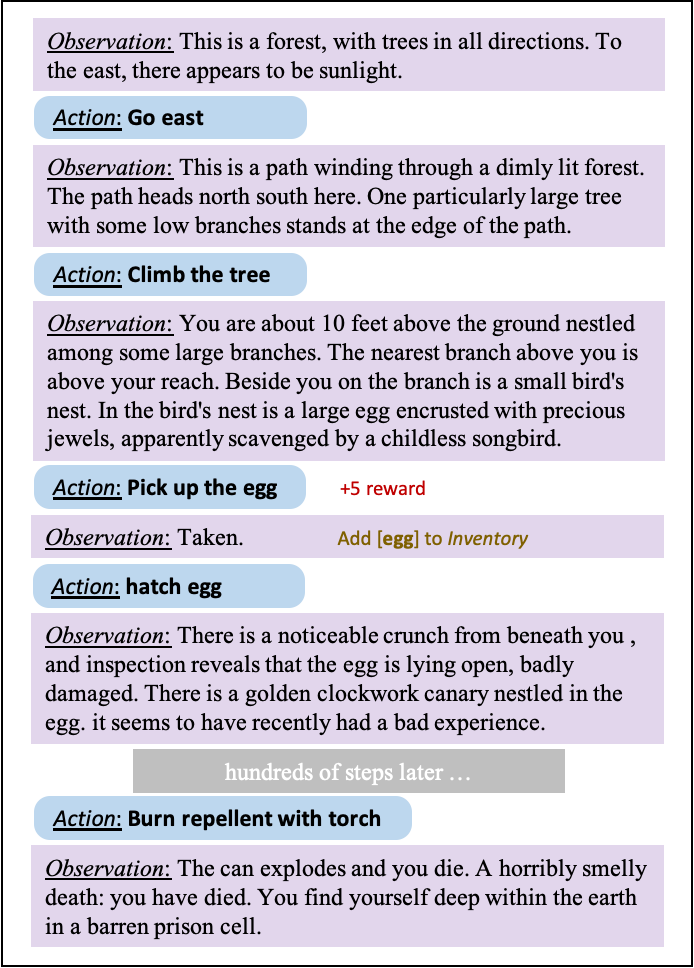}
\caption{\small{Classic dungeon game \textit{Zork1} gameplay sample. The player receives textual observations describing the current game state and sends textual action commands to control the protagonist. Various commonsense reasoning is illustrated in the texts of observations and commands from the gameplay interaction, such as spatial relations, objective manipulation, and physical relations.}}
\label{fig:game}
\vspace{-0.1in}
\end{figure}

\paragraph{Our Contributions}
To overcome these shortcomings, we derive \emph{commonsense reasoning tasks from the model-based reinforcement learning challenge of text games}. Our work is inspired by recent advances in interactive fiction (IF) game playing~\cite{hausknecht2019interactive,ammanabrolu2020graph,guo2020interactive}.
Figure~\ref{fig:game} illustrates sample gameplay of the classic game \emph{Zork1}.

The research community has recognized several commonsense reasoning problems in IF game playing~\cite{hausknecht2019interactive}, such as detecting valid actions and predicting the effects of different actions. In this work, we derive a commonsense evaluation related to the latter problem, i.e., predicting which is the most likely observation when applying an action to a game state.

Our approach of commonsense benchmark construction has several advantages. Specifically, it naturally relaxes the restrictions in commonsense types and reasoning forms.
First, we relax the limitation in commonsense types by noticing that predicting the next observation naturally requires various commonsense knowledge and reasoning types.
As shown in Figure~\ref{fig:game}, a primary commonsense type is spatial reasoning, e.g., \texttt{``climb the tree''} makes the protagonist up on a tree. 
Another primary type is reasoning with object interactions, such as with relationships, like keys can open locks; with object's properties, such as \texttt{``hatch egg''} will reveal ``things'' inside the egg; with physical reasoning, like \texttt{``burn repellent with torch''} leads to an explosion and kills the player.
The above interactions are much more comprehensive than the relationships defined in ConceptNet as used in previous datasets.
Second, we enforce the task to have more commonsense reasoning steps over simple commonsense validation. A large portion of IF game observations are narrative, and the next observation is less likely to be a sole statement of the action effect, but an extended narrates about what happens because of the effect.\footnote{For some actions, like get and drop objects, the returns are simple statements. We removed some of these actions. Details can be found in Section~\ref{sec:dataset}.}

Our benchmark designs based on the IF games support automatic data generation from multiple genres and domains, including dungeon crawl, Sci-Fi, mystery, comedy, and horror. From an RL perspective, our commonsense reasoning task formulation shares the essence of dynamics model learning for model-based RL solutions, like works related to world models and MuZero~\cite{ha2018world,schrittwieser2019mastering}. As a result, models developed on our benchmarks provide values to both commonsense reasoning and model-based reinforcement learning.

We introduce a new commonsense reasoning benchmark from four IF games in the \emph{Zork Universe}, {the \underline{Z}ork \underline{U}niverse \underline{C}ommonsense \underline{C}omprehension task} (\textbf{ZUCC}). Our experiments show that existing standard models perform poorly on the resulted benchmark, with a significant human-machine gap.

\section{ZUCC Dataset Construction}
\label{sec:dataset}

We pick games from the \emph{Zork Universe} that are supported by the \emph{Jericho} environment~\cite{hausknecht2019interactive}, namely \emph{zork1}, \emph{zork3}, \emph{enchanter}, \emph{sorcerer},\footnote{There is an excluded game, \emph{spellbreaker}, in Jericho that belongs to the Zork Universe. As the study in their paper shows, the game contains a large portion of non-standard actions that are usages of spells, and handling its non-standard vocabulary is beyond this paper's scope.} to construct our \dataname\ dataset.
This section first reviews the necessary definitions in the IF game domain; then describes how we construct our \dataname\ dataset as a forward prediction from the game walkthrough trajectories.

\subsection{Interactive Fiction Game Background}

\paragraph{Textual Observations and the POMDP Formulation}
An IF game-playing agent interacts with the game engine in multiple turns until the game is over or the maximum number of steps is reached. At the $t$-th turn, the agent receives a \emph{textual observation} describing the current game state $o_{t} \in O$ and an additional reward scalar $r_{t}$ indicating the game progress and it sends back a textual command $a_{t} \in A$ to control the protagonist. 

\paragraph{Trajectories and Walkthroughs}
A \emph{trajectory} in text game playing is a sequence of tuples $(o_t, a_{t}, r_{t}, o_{t+1})$ starting with the initial observation $o_0$. We define the \emph{walkthrough} of a text game as a trajectory that completes the game progress.

\subsection{Data Construction from the Forward Prediction Task}

\paragraph{The Forward Prediction Task}
We represent our commonsense reasoning benchmark as a next-observation prediction task, given the current observation and action. The benchmark construction starts with all the tuples in a walkthrough trajectory, and we then extend the tuple set by including all valid actions and their corresponding next-observations conditioned on the current observations in the walkthrough. Specifically, for a walkthrough tuple $(o_t, a_{t}, r_{t}, o_{t+1}, )$, we first obtain the complete valid action set $A_{t}$ for $o_t$. We sample and collect one next observation $o^j_{t+1}$ after executing the corresponding action $a^{j}_{t} \in A_t$. The next-observation prediction task is thus to select the next observation $o^j_{t+1}$ given $(o_t, a^j_{t})$ from the complete set of next observations $O_{t+1} = \{o^k_{t+1}, \forall k \}$. 

\paragraph{Data Processing}
We collect tuples from the walkthrough data provided by the Jericho environments. 
We detect the valid actions via the Jericho API and the game-specific templates.
Following previous work~\cite{hausknecht2019interactive}, we augmented the observation with the textual feedback returned by the command [$\textit{inventory}$] and [$\textit{look}$]. The former returns the protagonist's objects, and the latter returns the current location description.
When multiple actions lead to the same next-observation, we randomly keep one action and next-observation in our dataset.
We leave all the tuples from the \emph{zork3} game for evaluation.
We split the walkthrough of \emph{zork3}, keeping the first 136 tuples as a development set and the rest 135 tuples as a test set. 
We remove the \texttt{drop \textit{OBJ}} actions since it only leads to synthetic observations with minimal variety.
For each step $t$, we keep at most 15 candidate observations in $O_t$ for the evaluation sets. When there are more than 15 candidates, we select the candidate that differs most from $o_t$ with Rouge-L measure~\cite{lin-2004-rouge}. 

Table~\ref{tab:stats} summarizes statistics of the resulted \dataname\ dataset. The number of tuples is much larger in the test set because there are actions that do not have the form of \texttt{drop \textit{OBJ}} but have the actual effects of dropping objects. Through the game playing process, more objects will be collected in the inventory at the later stages. The test data will be much easier as long as these non-standard drop actions have been recognized. A similar problem happens to actions like \texttt{burn repellent} that can be performed at every step once the object is in the inventory. To deal with such problems in the test set, we finally down-sample these biased actions to achieve similar distributions in development and test sets. We perform down-sampling with rule-based methods. The resulted final version of the test set is denoted as \emph{Test (final)} in the table.

\paragraph{Remark on Further Impacts}
Our benchmark design opens opportunities beyond commonsense evaluation. For example, the form of our tasks, compared to the relevant tasks of next-sentence generation, such as the SWAG~\cite{zellers2018swag} and LIGHT~\cite{urbanek2019learning}, introduces actions as intervention, thus encourage causal reasoning. Therefore it has a potential impact on causal knowledge acquisition.
On the other hand, the \emph{partial observability} nature of IF games makes $o_t$ and $a^j_t$ not sufficient for predicting $o^j_t$ sometimes. Therefore our task encourages the development of structured abstract representations to summarize the history~\cite{ammanabrolu2020graph,ammanabrolu2020avoid}.

\begin{table*}[ht!]
    \small
    \centering
        \begin{tabular}{lcccc}
        \toprule
        \bf Sets & \bf \#WT Tuples  & \bf \#Tuples before Proc & \bf \#Tuples after Proc     \\
        \midrule
        Train & 913 & 17,741 & 10,498 \\
        Dev & 136  & 1,982 & 1,276  \\
        Test before down-sampling & 135 & 2,087 & 1,573 \\
        Test (final)  & - & - & 822  \\
        \bottomrule
        \end{tabular}
    \caption{
        Data statistics of our \dataname\ task. \textbf{WT} is short for walkthrough. Train set is from the game \textit{Zork1}, \textit{Enchanter}, and \textit{Sorcerer}. Both dev and test sets are from the game \textit{Zork3}. \label{tab:stats}}
\end{table*}

\section{Experiments}
We first benchmark the state-of-the-art models for natural language inference on our dataset, both with and without pre-trained Language Models (LMs). Then we conduct a human study on a sub-set of our development data to quantitatively measure the human performance and the human-machine gap.

\begin{table}[ht!]
    \small
    \centering
        \begin{tabular}{lcc}
        \toprule
        \bf Method & \bf Dev Acc & \bf Test Acc        \\
        \midrule
        Random Guess & 11.93 &9.39 \\
        \midrule
        Match LSTM & 57.52 & 62.17 \\
        BERT-siamese & 49.29 & 53.77   \\
        BERT-concat& 64.73 & 64.48  \\
        \midrule
        Human Average Performance* & 86.80 & -\\
        Human Expert Performance* & 96.40 & - \\
        \bottomrule
    \end{tabular}
    \caption{
    \label{tab:model_performance} Evaluation on our \dataname\ data. Human performance (*) is computed on a subset of data.}
    \label{tab:statistics}
\end{table}

\paragraph{Baselines} We compare the following baselines on the \dataname\ dataset.
In the model descriptions, the notations $o_t$, $a_t$ of observations and actions represent their word sequences.

\noindent$\bullet$ \textbf{Match LSTM} 
The neural attention model was proposed in~\cite{wang2016machine}, commonly used in natural language inference as baselines. Specifically, we concatenate $o_t$ and $a_t$ separated by a special split token as the premise and use the $o^j_{t+1}$ as the hypothesis. The matching scores for all $o^j_{t+1}$ are then fed to a softmax layer for the final prediction.

\noindent$\bullet$ \textbf{BERT Siamese}
The Siamese model uses a pre-trained BERT model to encode the current observation-action pair $(o_t, a_t)$ and candidate observations $\Tilde{o}^j_{t+1}, j = 1, ..., N$. All inputs to BERT start with the ``[CLS]'' token. $o_t$ and $a_t$ are concatenated by a ``[SEP]'' token:
\begin{align}
    \bm{h}_{t} &= \textrm{BERT}([o_t, a_t]),\quad \Tilde{\bm{h}}^j_{t+1} = \textrm{BERT}(\Tilde{o}^j_{t+1}),\nonumber\\
    \quad l_j &= f([\bm{h}_{t}, \Tilde{\bm{h}}^j_{t+1}, \bm{h}_{t} - \Tilde{\bm{h}}^j_{t+1}, \bm{h}_{t} * \Tilde{\bm{h}}^j_{t+1}]), \nonumber 
\end{align}
where $[\cdot ,\cdot ]$ denotes concatenation. $\bm{h}_{t}$ and $\Tilde{\bm{h}}^j_{t+1}$ are last layer hidden state vectors that correspond to the ``[CLS]'' token. 
Each candidate next-observation is scored by an output function $f$, and the logits are normalized by the $\textrm{softmax}$ function. We use the cross-entropy loss as the training objective. 

\noindent$\bullet$ \textbf{BERT Concat}
It represents the standard pairwise prediction mode of BERT. We concatenate $o_t$ and $a_t$ with a special split token as the first segment and treat $\Tilde{o}^j_{t+1}$ as the second. We then concatenate the two with the ``[SEP]'' token. We have a matching score for each $o^j_{t+1}$ with a linear mapping from the hidden state of the ``[CLS]'' token, and then feed the scores to a softmax layer for the final prediction.
This model is much less efficient than the former two; thus, it is not practical in IF game playing. Here we report its results for reference.

\paragraph{Implementation Details}
We experimented with training the three baselines on both full training tuples (biased training) and the processed training set (de-biased training). We reported the best development set performance for each model.

\paragraph{Results}
Table~\ref{tab:model_performance} summarizes the model performance. All three baselines manage to learn decent models, i.e., significantly better than a random guess.
For both Match LSTM and BERT-Siamese, the best development performance was found with de-biased training because this training setting is more consistent with the evaluation scenarios.

There is an exception for the BERT-Concat because the model is not learning in the de-biased training setting, i.e., the training accuracy stays around 10\%, the level of a random guess.
A possible reason is that the BERT-Concat model works directly on a complicated concatenated string of multiple types of inputs. Therefore it is challenging for it to distinguish the structures of input/output observations and actions. For example, it may not learn which parts of the inputs correspond to inventories.
To make the model work, we first pre-train the BERT-concat model on the biased training data until converging, then fine-tune the model on the de-biased data. This procedure finally gives the best performance on our \dataname.

Although the baselines are making progress, as shown in our human evaluation, the best development accuracy (64.73\%) is still far from human-level performance.
Compared to the human expert's near-perfect performance, the substantial performance gaps confirms that our \dataname\ captures challenging commonsense understanding problems.

\paragraph{Human Evaluation}
We present to the human evaluator each time a batch of tuples starting from the same observation $o_t$, together with its shuffled valid actions $A_{t+1}$ and next observations $O_{t+1}$.
The evaluators are asked to read the start observation $o_t$ first, then to align each $o \in O_{t+1}$ with an action $a \in A_{t+1}$.
Besides, for each observation $o$, besides guessing the action's alignment, the subjects are asked to answer a secondary question: whether the provided $o_t, o$ pair is sufficient for them to predict the action. 
If they believe there are not enough clues and their action prediction is based on a random guess, they are instructed to answer ``UNK'' to the second question.

We collect two sets of human predictions on 250 samples. The first set is annotated by one of the co-authors who have experience in interactive fiction game playing (but have \textbf{not} played \emph{Zork3} before). We denote the corresponding result as \emph{Human Expert Performance}. The second set is annotated by three of our co-authors who have never played IF games. The corresponding result is denoted as \emph{Human Average Performance}.
The corresponding accuracy is shown in Table~\ref{tab:model_performance}. The human expert performs more than 30\% higher compared to the machines. 
It is also interesting to see that even the human annotators who do not play IF games much can outperform the machine with more than 20\%. Since these annotators have not been trained for this task, their performance could represent human-level domain transferability with commonsense knowledge.

Finally, the annotators recognized 10.0\% cases with insufficient clues, indicating an upper-bound of methods without access to history observations.\footnote{Humans can still make a correct prediction by first eliminating most irrelevant options and then making a random guess.}

\section{Conclusion}
Interactive Fiction (IF) games encode plentiful and diverse commonsense knowledge of the physical world. In this work, we derive a commonsense reasoning benchmark \textbf{ZUCC} from IF games in the \textit{Zork Universe}. Taking the form of predicting the most likely observation when applying an action to a game state, our automatically generated benchmark covers comprehensive commonsense reasoning types such as spatial reasoning and object interaction, etc.  Our experiments on \textbf{ZUCC} show that current popular neural models have limited performance compared to humans. 

\bibliographystyle{acl_natbib}
\bibliography{emnlp2020}

\end{document}